\documentclass[11pt]{article}
\usepackage{coling2020}
\usepackage{times}
\usepackage{url}
\usepackage{latexsym}
\usepackage{comment}
\usepackage{graphicx}
\usepackage[multiple]{footmisc}
\usepackage{multirow}
\usepackage{amsmath}
\usepackage{booktabs}
\usepackage{arydshln}
\usepackage{tcolorbox}
\usepackage{dingbat}
\usepackage{adjustbox}
\usepackage{rotating}
\title{\center
SemEval-2020 Task 9: Overview of Sentiment Analysis of \\
Code-Mixed Tweets
}

\author{
Parth Patwa\textsuperscript{1$^*$} \quad 
Gustavo Aguilar\textsuperscript{2$^*$} \quad
Sudipta Kar\textsuperscript{2}\quad 
Suraj Pandey\textsuperscript{3}  \\ 
\bf Srinivas PYKL\textsuperscript{1} \quad
\bf Bj{\"o}rn Gamb{\"a}ck\textsuperscript{4}\quad
\bf Tanmoy Chakraborty\textsuperscript{3}\\ 
\bf Thamar Solorio\textsuperscript{2}\quad
\bf Amitava Das\textsuperscript{5} \\
\textsuperscript{1}IIIT Sri City, India \quad
\textsuperscript{2}University of Houston, TX \quad
\textsuperscript{3}IIIT Delhi, India \\
\textsuperscript{4}NTNU, Norway \quad
\textsuperscript{5}Wipro AI Labs, India\\
\tt \textsuperscript{1}\{parthprasad.p17, srinivas.p\}@iiits.in\\
\tt \textsuperscript{2}\{gaguilaralas, skar3, tsolorio\}@uh.edu\\
\tt \textsuperscript{3}\{suraj18025, tanmoy\}@iiitd.ac.in \\
\tt\textsuperscript{4}{gamback@ntnu.no} 
\tt\textsuperscript{5}amitava.das2@wipro.com
}

\date{}

\begin{document}
\maketitle
\renewcommand{\thefootnote}{\fnsymbol{footnote}}
\footnotetext[1]{Equal contribution.}
\renewcommand*{\thefootnote}{\arabic{footnote}}
\setcounter{footnote}{0}

\begin{abstract}
In this paper, we present the results of the SemEval-2020 Task 9 on Sentiment Analysis of Code-Mixed Tweets (SentiMix 2020).\footnote{\url{https://ritual-uh.github.io/sentimix2020/}}
We also release and describe our Hinglish (Hindi-English) and Spanglish (Spanish-English) corpora annotated with word-level language identification and sentence-level sentiment labels. These corpora are comprised of 20K and 19K examples, respectively. The sentiment labels are - Positive, Negative, and Neutral. SentiMix attracted 89 submissions in total including 61 teams that participated in the Hinglish contest and 28 submitted systems to the Spanglish competition.
The best performance achieved was 75.0\% F1 score for Hinglish and 80.6\% F1 for Spanglish.
We observe that BERT-like models and ensemble methods are the most common and successful approaches among the participants.
\end{abstract}

\blfootnote{
    %
    %
    %
    
     
    
    \vspace{0.65cm}  
    \hspace{-0.65cm}  
    This work is licensed under a Creative Commons 
    Attribution 4.0 International License.
    License details:
    \url{http://creativecommons.org/licenses/by/4.0/}.
}

%
\section{Introduction}\label{intro}


The evolution of social media texts such as blogs, micro-blogs (e.g., Twitter), and chats (e.g., WhatsApp and Facebook messages) has created many new opportunities for information access and language technologies. However, it has also posed many new challenges making it one of the current prime research areas in Natural Language Processing (NLP). 

Current language technologies primarily focus on English \cite{young2020}, yet social media platforms demand methods that can also process other languages as they are inherently multilingual environments.\footnote{Statistics show that half of the messages on Twitter are in a language other than English \cite{Schroeder:2010}.}
Besides, multilingual communities around the world regularly express their thoughts in social media employing and alternating different languages in the same utterance. 
This mixing of languages, also known as code-mixing or code-switching,\footnote{We use code-mixing and code-switching interchangeably.} is a norm in multilingual societies and is one of the many NLP challenges that social media has facilitated. 

\subsection{Code-Mixing Challenges}

In addition to the writing aspects in social media, such as flexible grammar, permissive spelling, arbitrary punctuation, slang, and informal abbreviations \cite{baldwin-etal-2015-shared,eisenstein-2013-phonological}, code-mixing has introduced a diverse set of linguistic challenges.
For instance, multilingual speakers tend to code-mix using a single alphabet regardless of whether the languages involved belong to different writing systems (i.e., language scripts). This behavior is known as transliteration, and code-mixers rely on the phonetic patterns of their writing (i.e., the actual sound) to convey their thoughts in the foreign language (i.e., the language adapted to a new script) \cite{sitaram2019survey}. 
Another common pattern in code-mixing is the alternation of languages at the word level. This behavior often happens by inflecting words from one language with the rules of another language \cite{SolorioAndLiu:08b}. 
For instance, in the second example below, the word \textit{push\underline{es}} is the result of conjugating the English verb \textit{push} according to Spanish grammar rules for the present tense in third person (in this case, the inflection \textit{\underline{-es}}). The
Hinglish example shows that phonetic Latin script typing is a popular practice in India, instead of using Devanagari script to write Hindi words.
We capture both transliteration and word-level code-mixing inflections in the Hinglish and Spanglish corpora of this competition, respectively.

\begin{quote}
\centering
\textit{
    Aye\textsubscript{\texttt{HI}} 
    aur\textsubscript{\texttt{HI}} 
    enjoy\textsubscript{\texttt{EN}} 
    kare\textsubscript{\texttt{HI}}
}\\
\textbf{Eng. Trans.: }  come and enjoy
\\[.5ex]
\centering
\textit{
    No\textsubscript{\texttt{SP}} 
    me\textsubscript{\texttt{SP}} 
    \textbf{push\textit{es}}\textsubscript{\texttt{EN}} 
    please\textsubscript{\texttt{EN}}
}
\\
\textbf{Eng. Trans.: }  Don't push me, please\\
\end{quote}

Considering the previous challenges, code-mixing demands new research methods where the focus goes beyond simply combining monolingual resources to address this linguistic phenomenon.
Code-mixing poses difficulties in a variety of language pairs and on multiple tasks along the NLP stack, such as word-level language identification, part-of-speech tagging, dependency parsing, machine translation, and semantic processing \cite{sitaram2019survey}. 
Conventional NLP systems heavily rely on monolingual resources to address code-mixed text, limiting them when properly handling issues such as phonetic typing and word-level code-mixing. 

\subsection{Code-Mixing as a Global Linguistic Phenomenon}

Naturally, code-mixing is more common in geographical regions with a high percentage of bi- or multilingual speakers, such as in Texas and California in the US, Hong Kong and Macao in China, many European and African countries, and the countries in South-East Asia. 
Multilingualism and code-mixing are also widespread in India, which has more than 400 languages \cite{ethnologue} with about 30 languages having more than 1 million speakers. 
Language diversity and dialect changes trigger Indians to frequently change and mix languages, particularly in speech and social media contexts. 
As of 2020, Hindi and Spanish have over 630 million and over 530 million speakers \cite{ethnologue}, respectively, ranking them in 3rd and 4th place based on the number of speakers worldwide, which speaks of the relevancy of using these languages in our code-mixing competition.

\subsection{SentiMix Overview}

This paper provides an overview of the SemEval-2020 Task 9 competition on sentiment analysis of code-mixed social media text (SentiMix). 
Specifically, we provide code-mixed text annotated with word-level language identification and sentence-level sentiment labels (negative, neutral, and positive). 
We release our Hinglish (Hindi-English) and Spanglish (Spanish-English) corpora, which are comprised of 20K and 19K tweets, respectively. We describe general statistics of the corpora as well as the baseline for the competition.

We received 61 final submissions for Hinglish and 28 for Spanglish, adding to a total number of 89 submissions. We received 33 system description papers.
We provide an overview of the participants' results and describe their methods at a high level. 
Notably, the majority of these methods employed BERT-like and ensemble models to reach competitive results, with the best performers reaching 75.0\% and 80.6\% F1 scores for Hinglish and Spanglish on held-out test data, respectively.
We hope that this shared task will continue to catch the NLP community's attention on the linguistic code-mixing phenomenon.

\section{Related Work }\label{sec:related_work}

 
Linguists ~\cite{VERMA1976153,BOKAMBA198821,singh_1985} studied the phenomena of code-mixing and intra-sentential code-switching and found that processing code-mixed language is much more complicated than monolingual text.
Code-mixing is often found on social media which contains a lot of nonstandard spellings of words and unnecessary capitalization \cite{das-gamback-2014-identifying}, making the task more difficult. Naturally, the difficulty will increase as the amount of code-mixing increases. To quantify the level of code-switching between languages in a sentence, \newcite{gamback-das-2016-comparing} introduced a measure called Code Mixing Index (CMI) which considers the number of tokens of each language in a sentence and the number of tokens where the language switches. 

Finding the sentiment from code-mixed text has been attempted by some researchers. \newcite{mohammad-etal-2013-nrc} used SVM-based classifiers to detect sentiment in tweets and text messages using semantic information. 
\newcite{bojanowski-etal-2017-enriching} proposed a skip-gram based word representation model that classifies the sentiment of tweets and provides an extensive vocabulary list for language.
\newcite{GIATSOGLOU2017214} trained lexicon-based document vectors, word embedding, and hybrid systems with the polarity of words to classify the sentiment of a tweet. 
\newcite{sharma-etal-2016-shallow} attempted shallow parsing of code-mixed data obtained from online social media, and \newcite{chittaranjan-etal-2014-word} tried word-level identification of code-mixed data to classify the sentiment. Some researchers also tried normalizing the text with lexicon lookup for sentiment analysis of code-mixed data \cite{7275819}.

To advance research in code-mixed language processing, few workshops have also been conducted. Four successful series of Mixed Script Information Retrieval have been organized at the Forum for Information Retrieval Evaluation (FIRE) \cite{overview-mixed-script-information-retrieval-msir_2013,overview-mixed-script-information-retrieval-msir_2014,overview-fire-2015-shared-task-mixed-script-information-retrieval,overview-mixed-script-information-retrieval-msir}. Three workshops on Computational Approaches to Linguistic Code-Switching (CALCS) have been conducted which included shared tasks on language identification and Named Entity Recognition (NER) in code-mixed data \cite{solorio-etal-2014-overview,molina-etal-2016-overview,calcs2018shtask}.
For our SentiMix Spanglish dataset, we adopt the SentiStrength \cite{vilares-etal-2015-sentiment} annotation mechanism and conduct the annotation process over the unified corpus from the three CALCS workshops.


\section{Task Description}\label{sec:task_definition}
Although code-mixing has received some attention recently, properly annotated data is still scarce. We run a shared task to perform sentiment analysis of code-mixed tweets crawled from social media. Each tweet is classified into one of the three polarity classes - Positive, Negative, Neutral. Each tweet also has word-level language marking. We release two datasets - Spanglish and Hinglish. 

We used CodaLab\footnote{Hinglish: \url{https://competitions.codalab.org/competitions/20654}}\footnote{Spanglish: \url{https://competitions.codalab.org/competitions/20789}} to release the datasets and evaluate submissions. Initially, the participants had access only to train and validation data. They could check their system's performance on the validation set on a public leaderboard. Later, a previously unseen test set was released, and the performance on the test set was used to rank the participants. Only the first three submissions on the test set by each participant were considered, to avoid over-fitting on the test set. The ranking was done based on the best out of the three submissions. There was no distinction between constrained and unconstrained systems, but the participants were asked to report what additional resources they have used for each submitted run.

We release 20k labeled tweets for Hinglish and $\approx$ 19k labeled tweets for Spanglish. In both the datasets,\footnote{Both the datasets are available at \url{https://zenodo.org/record/3974927#.XyxAZCgzZPZ}} in addition to the tweet level sentiment label, each tweet also has a word-level language label. The detailed distribution is provided in Table \ref{tab:data_stat}. Some annotated examples are provided in Table \ref{tab:examples_word_sentiment}. Although this task focuses on sentiment analysis, the data has word-level language marking and can be used for other NLP tasks.

\subsection{Evaluation Metric}

To evaluate the performance and rank the participants, we use weighted F1 score on the test data, across the positives, negatives, and neutral examples. 
\begin{center}
     \text{F1} = \( \frac{2 \times \text{Precision} \times \text{Recall}}{ \text{Precision} + \text{Recall}} \)
\end{center}
where,
\begin{center}
     \text{Precision} = \( \frac{\text{True Positives}} {\text{True Positives} + \text{False Positives}} \) , 
     \text{Recall} = \( \frac{\text{True Positives}}{\text{True Positives} + \text{False Negatives}} \)
\end{center}

The F1 scores are calculated for each class and then their average is weighted by support (number of true instances for each class). We use a weighted F1 score since the number of instances per class is not equal. Other than the F1 score, we also calculate precision and recall for each class to analyze and have a better understanding of \textit{false positives} and \textit{false negatives}.


\section{Dataset}
The datasets consist of tweets labeled into one of the three classes: 
\begin{itemize}
    \item \textbf{Positive (Pos)}: 
    Tweets which express happiness, praise a person, group, country or a product, or applaud something. Hinglish example:
    \textit{ ``bholy bhayaa. Ufffff dil jeet liya ap ne. Love you imran bhai. Mind blowing ap ki acting hai.'' } 
    (bholy bhayaa, you won hearts. love you imran bhai your acting is mind blowing).
    Spanglish example: 
    \textit{ ``We all here waiting pa ke juege mex :)''} 
    (We all here waiting for Mexico to play :)).
    
    \item \textbf{Negative (Neg)}: Tweets which attack a person, group, product or country, express disgust or unhappiness towards something, or criticize something. 
    Hinglish example: \textit{``You efficiency of anchoring a program is continuously deteriorating. Ab to dekhne ki himmat hi nahi''} (Your efficiency of anchoring is continuously deteriorating. Now can't even dare to watch it)
    Spanglish example: \textit{``Eres una cualkiera yes u are.''} (You are a tramp, yes you are.)
    
    \item \textbf{Neutral (Neu)}: Tweets which state facts, give news or are advertisements. In general those which don't fall into the above 2 categories. Hinglish example: \textit{``Nahi wo is news ko defend kerne ki koshesh ker rhe hain h''} (No, they are trying to defend this news).
    Spanglish example: \textit{``My phone looks ratchet todo crack''} (My phone looks ratchet all crack).
\end{itemize}

\begin{table}[!hbtp]
\centering
\begin{tabular}{|l|l|l|l|l|l|}
\hline
\textbf{Language}                   & \textbf{Split}      & \textbf{Total}   & \textbf{Positive}              & \textbf{Neutral}              & \textbf{Negative}     \\ \hline
\multirow{4}{*}{Hinglish}  & Train      & 14,000  & 4,634 (33.10\%)  & 5,264 (37.60\%)  & 4102 (29.30\%) \\ \cline{2-6} 
                           & Validation & 3,000   & 982 (32.73\%)    & 1,128 (37.60\%)  & 890 (29.67\%) \\ \cline{2-6} 
                           & Test       & 3,000   & 1,000 (33.33\%)  & 1,100 (36.67\%)  & 900 (30\%) \\ \cline{2-6} 
                           & \textbf{Total}      & 20,000  & 6,616 (33.08\%)  & 7,492 (37.46\%)  & 5892 (29.46\%) \\ 
                           \hline
\multirow{4}{*}{Spanglish} & Train      & 12,002  & 6,005 (50.03\%)  & 3,974 (33.11\%)  & 2,023 (16.85\%)\\ \cline{2-6} 
                           & Validation & 2,998   & 1,498 (49.96\%)  & 994 (33.15\%)    & 506 (16.87\%) \\ \cline{2-6} 
                           & Test       & 3,789   & 3,061 (80.78\%)  & 206 (5.43\%)     & 522 (13.77\%) \\ \cline{2-6} 
                           & \textbf{Total}      & 18,789  & 10,564 (56.22\%) & 5,174 (27.53\%)  & 3,051 (16.23\%) \\ \hline
\end{tabular}
    \caption{Class-wise statistics of the dataset for train, validation, and test set. We put special care to make a balanced class-wise distribution for Hinglish.}
    \label{tab:data_stat}
\end{table}

Both the Hinglish and Spanglish datasets are released using the previous sentiment label scheme. However, each dataset has been annotated separately as the studies were independent before the organization of this competition. We provide the data collection and annotation details in the following subsections.

\subsection{Hinglish}
\paragraph{Data Collection:}
First, we make a list of all the Hindi tokens from the dataset provided by \cite{das_2017-data}. From that list, we remove those tokens which are common to Hindi and English (example '\textit{the}' can be used in both the languages). Then we use Twitter API \footnotemark[7] to crawl those tweets from twitter which have at least one word from the list. The list has 10786 tokens. Some words from the list are: \textit{kuch, tu, gaya, raha, aaj, apne, tum, gaye, sath} etc.

\paragraph{Language and Sentiment Annotation:} For word-level language marking we use an automated tool 
released by \newcite{10.1145/2824864.2824872}. The tokens are labeled into HIN - Hindi, ENG - English, or O - other. 
For tweet level sentiment labels, we took the help of around 60 annotators who were bilingual/multilingual, proficient in Hindi and had Hindi as their first or second language. Each tweet was shown to two annotators, and it was selected if their annotations matched, else it was discarded. They used a simple website designed for this purpose to annotate the data. Each tweet was shown on a page that had a radio button for each label. The annotators first had to enter their unique id, then they could either select a sentiment option for a tweet and send or choose to skip the tweet. 

\paragraph{Statistics:} Table \ref{tab:data_stat} gives detailed class-wise distribution of the tweets. Although Neutral is the majority class for Hinglish, the dataset is not too imbalanced. The class-wise distribution is similar for all three splits.  Table \ref{tab:examples_word_sentiment} shows some examples of tweets marked with language and sentiment tags. 
The average CMI for Hinglish train, validation, and test set is  25.32, 25.53, and 25.13 respectively. The inter-annotator agreement is 55\%.

\subsection{Spanglish}
\label{sec:spang}
\paragraph{Data Collection:} 
We use the Spanish-English data from the CALCS
workshops \cite{W14-3907,molina-etal-2016-overview,calcs2018shtask}. 
In the first workshop \cite{W14-3907}, the data was collected by crawling tweets from specific locations with a strong presence of Spanish and English speakers (e.g., California and Texas). The collection process was conducted using common words from each language through the Twitter API.\footnote{\url{https://developer.twitter.com/}}
In the second workshop \cite{molina-etal-2016-overview}, the organizers provided a new test set collected with a more elaborated process. They selected big cities where bilingual speakers are common (e.g., New York and Miami). Then, they localized Spanish radio stations that showed code-mixed tweets. Such radio stations led to users that also practice code-mixing. 
Similar to the third workshop \cite{calcs2018shtask}, we take the CALCS data and extend it for sentiment analysis. 
It is worth noting that a large number of tweets in the corpora only contain monolingual text (i.e., no code-mixing). 
Considering that, and after merging the two corpora, we prioritize the tweets that show code-mixed text to build the SentiMix corpus. 
We ended up incorporating 280 monolingual tweets per language (English, Spanish) in the test set.

\paragraph{Annotation:} Since we use the data from the previous CALCS workshops, we did not need to undergo the token-level annotation process for language identification (LID). 
We adopted the CALCS LID label scheme, which is comprised of the following eight classes:
\texttt{lang1} (English), 
\texttt{lang2} (Spanish),
\texttt{mixed} (partially in both languages), 
\texttt{ambiguous} (either one or the other language), 
\texttt{fw} (a language different than \texttt{lang1} and \texttt{lang2}), 
\texttt{ne} (named entities), 
\texttt{other}, and 
\texttt{unk} (unrecognizable words). 

\begin{figure}[h!]
    \centering
    {%
    \setlength{\fboxsep}{8pt}%
    \setlength{\fboxrule}{1pt}%
    \fbox{\includegraphics[width=0.8\linewidth]{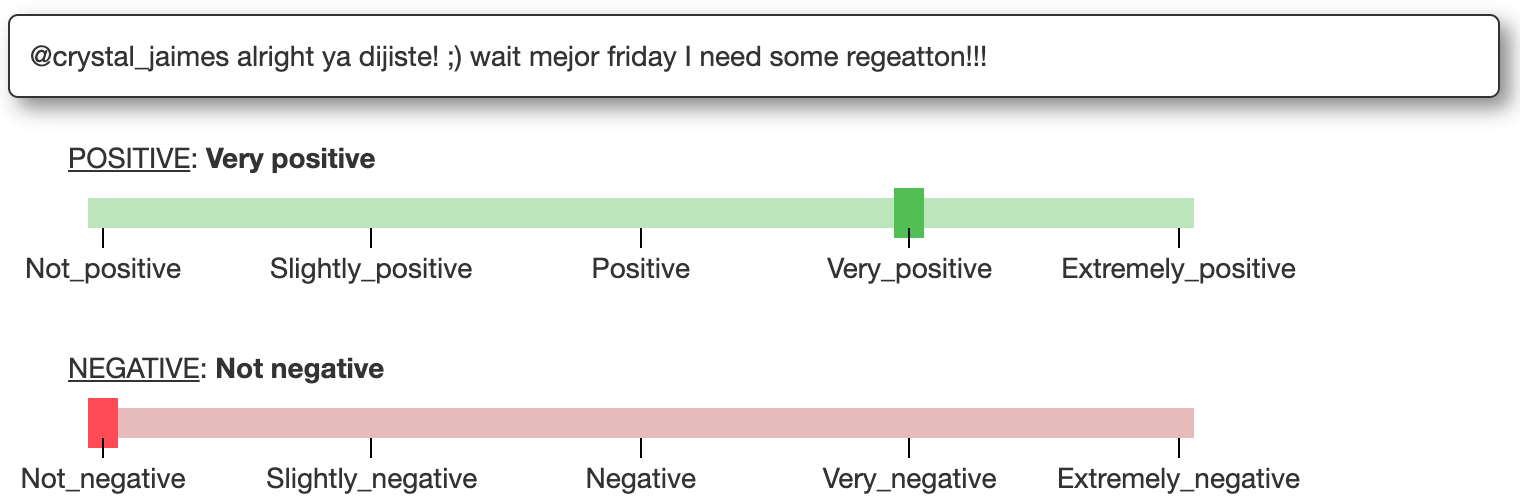}}%
    }%
    \caption{A screenshot of the Spanglish annotation interface.}
    \label{fig:spa_eng_annotation_example}
\end{figure}

For the annotations of the sentiment labels, we follow the SentiStrength\footnote{\url{http://sentistrength.wlv.ac.uk/}} strategy \cite{thelwall2010sentiment,vilares-etal-2015-sentiment}. 
That is, we provide positive and negative sliders to the annotators. 
Each slider denotes the strength for the corresponding sentiment, and the annotators can choose the level of the sentiment they perceived from the text (see Figure \ref{fig:spa_eng_annotation_example}). 
The range of the sliders is discrete and included strengths from 1 to 5 with 1 being \textit{no strength} (i.e., no positive or negative sentiment) and 5 the strongest level.
Using two independent sliders allowed the annotators to process the positive and negative signals without excluding one from the other, letting them provide mixed sentiments for the given text \cite{10.3389/fpsyg.2015.00428}. 
Once the sentiment strengths were specified, we converted them into a 3-way sentiment scale (i.e., \texttt{positive}, \texttt{negative}, and \texttt{neutral}). 
We simply subtract the negative strength from the positive strength, and mark the text as \texttt{positive} if the result was greater than zero, \texttt{negative} if less than zero, or \texttt{neutral} otherwise.

We annotate each tweet with the help of three annotators from Amazon Mechanical Turk.\footnote{\url{https://requester.mturk.com/}} 
We regulate the annotations by using quality questions within every assignment\footnote{An assignment is done by a single annotator.} of a HIT (Human Intelligence Task). 
Every assignment has ten tweets, two of them were for quality control (i.e., the annotation was already known) and the other eight tweets were the ones to annotate.\footnote{We use the assignment review policy \texttt{ScoreMyKnownAnswers/2011-09-01}.} 
The annotators had to have at least one quality control tweet right so that the assignment (i.e., the ten tweets) was not automatically rejected.
Since the sentiment analysis task is arguably arbitrary, we provided multiple valid levels of strength for the quality control tweets.
If an assignment was rejected, then another annotator was automatically required to complete the HIT until three annotations were accepted.
Also, we automatically approved HITs if their 3-way sentiment inter-annotator agreement was over 66\%.\footnote{We use the HIT review policy \texttt{SimplePlurality/2011-09-01}.} 
Otherwise, we evaluated manually the annotations and decide whether to extend the assignments or mark the sentiment labels ourselves for the trivial cases. 
After merging the annotations, we gave a pass over the data and manually corrected annotations that were unambiguously wrong.

\paragraph{Statistics:} 
The Spanglish class-level distribution of the partitions appear in Table \ref{tab:data_stat}. 
Notably, the data is highly imbalanced towards the positive class covering about 56\% in the entire Spanglish corpus, while the negative and neutral classes account for around 16\% and 27\%, respectively. 
The reason for this imbalance distribution is that we did not collect the data following a sentiment-oriented crawling strategy (e.g., searching by sentiment-related keywords).
Instead, we just extended the original corpus, which happens to be mostly positive. 
The intention to proceed in this way is to enrich the original corpus annotations with sentiment-level labels.
Moreover, the splits do not share the same distribution (i.e., development and test are more skewed than training) because we were annotating data on-demand rather than having available the entire corpus at any stage of the competition.
Some annotated examples are provided in Table \ref{tab:examples_word_sentiment}.
The average CMI for the train, validation, and test sets are 21.84, 20.52, and 17.23, respectively. 


\begin{table}[!htbp]
\centering
\begin{tabular}{|l|l|l|}
\hline
Language &
  Tweet &
  Class \\ \hline
Spanglish &
  \begin{tabular}[c]{@{}l@{}}
    @username\textsubscript{\texttt{other}}
    ha\textsubscript{\texttt{lang2}}
    pos\textsubscript{\texttt{ambiguous}}
    have\textsubscript{\texttt{lang1}}
    fun\textsubscript{\texttt{lang1}}
    its\textsubscript{\texttt{lang1}} \\ 
    pretty\textsubscript{\texttt{lang1}}
    te\textsubscript{\texttt{lang2}} 
    subes\textsubscript{\texttt{lang2}}
    al\textsubscript{\texttt{lang2}}
    horse\textsubscript{\texttt{lang1}}
    its\textsubscript{\texttt{lang1}} 
    cute\textsubscript{\texttt{lang1}} \\
    lol\textsubscript{\texttt{lang1}} 
    \\
    (@username ah, then have fun, it's pretty, you ride the horse, it's cute lol)
  \end{tabular} &
  Positive \\ \hline
Spanglish &
  \begin{tabular}[c]{@{}l@{}}
    Cuando\textsubscript{\texttt{lang2}}
    Mis\textsubscript{\texttt{lang2}}
    parents\textsubscript{\texttt{lang1}}
    me\textsubscript{\texttt{lang2}}
    dejan\textsubscript{\texttt{lang2}}
    ir\textsubscript{\texttt{lang2}}
    el\textsubscript{\texttt{lang2}} \\
    date\textsubscript{\texttt{lang1}}
    me\textsubscript{\texttt{ambiguous}} 
    Keda\textsubscript{\texttt{lang2}}
    Mal\textsubscript{\texttt{lang2}}
    /\textsubscript{\texttt{other}}
    .\textsubscript{\texttt{other}}
    -\textsubscript{\texttt{other}} \\
    No\textsubscript{\texttt{lang2}}
    MAMEN\textsubscript{\texttt{lang2}}
     \\ 
     (When my parents let me go, my date is cancelled / . - You're kidding me!)
  \end{tabular} &
  Negative \\ \hline
Spanglish &
  \begin{tabular}[c]{@{}l@{}}
    Tengo\textsubscript{\texttt{lang2}}
    hungry\textsubscript{\texttt{lang1}}
    mhm\textsubscript{\texttt{unk}} 
    \\
    (I'm hungry mhm)
  \end{tabular} &
  Neutral \\ \hline
Hinglish &
  \begin{tabular}[c]{@{}l@{}}Congratulations\textsubscript{\texttt{ENG}} Sir\textsubscript{\texttt{ENG}} we\textsubscript{\texttt{ENG}} proud\textsubscript{\texttt{ENG}} of\textsubscript{\texttt{ENG}} you\textsubscript{\texttt{ENG}} ..\textsubscript{\texttt{O}} Aap\textsubscript{\texttt{HIN}}\\ pr\textsubscript{\texttt{HIN}} pura\textsubscript{\texttt{HIN}} jakeen\textsubscript{\texttt{HIN}} hai\textsubscript{\texttt{HIN}} ..\textsubscript{\texttt{O}} aap\textsubscript{\texttt{HIN}} bohat\textsubscript{\texttt{HIN}} achaa\textsubscript{\texttt{HIN}}n home\textsubscript{\texttt{HIN}} \\  minister\textsubscript{\texttt{ENG}} Honga\textsubscript{\texttt{HIN}} ..\textsubscript{\texttt{O}} )\textsubscript{\texttt{O}}\\ (Congratulations sir we are proud of you.. We believe in you..  \\ You will be a very good home minister.. )\end{tabular} &
  Positive \\ \hline
Hinglsih &
  \begin{tabular}[c]{@{}l@{}}Hostelite\textsubscript{\texttt{ENG}} k\textsubscript{\texttt{ENG}} naam\textsubscript{\texttt{HIN}} pe\textsubscript{\texttt{HIN}} dhabba\textsubscript{\texttt{HIN}}  ho\textsubscript{\texttt{HIN}} tum\textsubscript{\texttt{HIN}}\\ (you are a blot on the name of a hostelite)\end{tabular} &
  Negative \\ \hline
Hinglish &
  \begin{tabular}[c]{@{}l@{}}Warm\textsubscript{\texttt{ENG}} up\textsubscript{\texttt{ENG}} match\textsubscript{\texttt{ENG}} to\textsubscript{\texttt{ENG}} theek\textsubscript{\texttt{HIN}} thaak\textsubscript{\texttt{HIN}} chal\textsubscript{\texttt{HIN}} ra\textsubscript{\texttt{HIN}} hai\textsubscript{\texttt{HIN}} \\ (Warm up match is going fine)\end{tabular} &
  Neutral \\ \hline
\end{tabular}
\caption{Examples of labeled tweets. Code-mixing often refers to the juxtaposition of linguistic units from two or more languages in a single conversation or sometimes even a single utterance. These examples emphasize on the fact that people don't do only phrase, or tag-mixing as it was a belief in the linguistic forum until now.}
     \label{tab:examples_word_sentiment}
\end{table}


\section{Baseline}\label{sec:baseline}
We develop our baseline system using the pre-trained \textit{multilingual} BERT (M-BERT; \newcite{devlin-etal-2019-bert}).
M-BERT was trained on 104 languages' entire Wikipedia dump and the WordPiece \cite{DBLP:journals/corr/WuSCLNMKCGMKSJL16} vocabulary of this model contains 110K sub-word tokens from these 104 languages.
To balance the risk of low-resource languages being under-represented or over-fitted due to small training resources during pre-training, exponentially smoothed weighting was performed on the data during pre-training data creation and vocabulary creation.
Although M-BERT was trained on monolingual data from different languages, it is capable of multilingual generalization in code-switching scenarios \cite{pires-etal-2019-multilingual}.

We use the Transformers \cite{Wolf2019HuggingFacesTS} library to implement our framework and we fine-tune the pre-trained \texttt{BERT-Base, Multilingual Cased} model separately for each of the two languages.
Based on our observation on the training split for each dataset, we set the highest sequence length to 40 and 56 tokens for Spanglish and Hinglish, respectively.
Then, we fine-tune the model for three epochs using AdamW \cite{loshchilov2018decoupled} optimizer ($\eta = 2e^{-5}$).
\section{Participation and Top Performing Systems}\label{sec:participants}

We received an overwhelming response for both Hinglish and Spanglish. 61 teams submitted their systems for Hinglish and 28 teams submitted their systems for Spanglish. 16 teams submitted to both Hinglish and Spanglish. We received 33 system description papers in total. The embeddings and techniques used by the participants are tabulated in Table \ref{tab:teams_techniques}. The team names, Codalab names, and their corresponding description papers are provided in Appendix (Table \ref{tab:participants_bib}). We provide a summary of the top teams below (Codalab usernames are mentioned in parentheses) : 

\textbf{Top  Hinglish Systems @ SentiMix}
\begin{itemize}
\item \textbf{KK2018 (kk2018) } used pre-trained XLM-R\cite{conneau2019unsupervised}  which was trained with 100 languages. They trained it with adversarial (intentionally designed to make model cause a mistake) examples. To create adversarial examples, they used the formula proposed by \cite{Miyato2016VirtualAT} where the perturbation is created using  the gradient of the loss function.
  
\item \textbf{MSR India (genius1237)} used embeddings from XLM-R as inputs to a classification layer. They also do so with multiligual BERT.

\item \textbf{Reed (gopalvinay)} Finetuned BERT and claimed that pre-training of BERT is not of much use. They also tried bag-of-words based feedforward networks. 

\item \textbf{BAKSA (ayushk)} used XLM-R\cite{conneau2019unsupervised}  multilingual embeddings ( a transformer-based masked language model trained on 100 languages) followed by ensemble model of CNN and self attention architecture.
\end{itemize}

\textbf{Top Spanglish Systems @ SentiMix}
\begin{itemize}
    
\item \textbf{XLP (LiangZhao)} augmented the data using machine translation. Then they used pre-trained embeddings made by Facebook Research (XLMs)\cite{lample2019cross} followed by CNN classifier of linear classifier (fully connected layer). They optimized a weighted loss function based on the complexity of code-mixing. 

\item \textbf{Voice@SRIB (asking28)} applied multiple pre-processing steps and used Ensemble model by combining CNN, self-attention and LSTM based model.

\item \textbf{	Palomino-Ochoa (dpalominop)} combined a transfer learning scheme based on ULMFit \cite{howard2018universal} with the-state-of-the-art language model BERT.

\item \textbf{ HPCC-YNU (kongjun)} used word and character embeddings as input to BiLSTM with attention.     
\end{itemize}

\section{Results and Analysis}\label{sec:results}

Table \ref{tab:hinglish_results} and Table \ref{tab:spanglish_results} show top 15 participants \footnote{ Results for all the participants are available at \url{https://ritual-uh.github.io/sentimix2020/}} for Hinglish and Spanglish respectively. For Hinglish the top 15 participants lie between 75\% and 68.6\% F1 score. The participants in the middle of the table are quite close to each other. 44 participants beat the baseline whereas 17 could not. For Spanglish, the top 15 F1 scores lie between 80.6\% and 71.0\% and most are in mid 70s. 22 teams were able to beat the baseline whereas 6 could not. The results are much better for positive than for other two classes due to the data imbalance.

\begin{table}[!htbp]
\centering
\resizebox{0.95\textwidth}{!}{

\begin{tabular}{cl  rrr @{\hspace{0.3in}} rrr  @{\hspace{0.3in}} rrr  @{\hspace{0.3in}} l}
\toprule

\textbf{Rank} & \textbf{System} & 
\multicolumn{3}{c @{\hspace{0.3in}}}{\textbf{Positive}} & 
\multicolumn{3}{c @{\hspace{0.3in}}}{\textbf{Neutral}} & 
\multicolumn{3}{c @{\hspace{0.3in}}}{\textbf{Negative}} & 
\textbf{Avg.} \\ \addlinespace[0.2cm]

& & \multicolumn{1}{c}{P} & \multicolumn{1}{c}{R} & \multicolumn{1}{c @{\hspace{0.3in}}}{F1} & 
\multicolumn{1}{c}{P} & \multicolumn{1}{c}{R} & \multicolumn{1}{c @{\hspace{0.3in}}}{F1} & 
\multicolumn{1}{c}{P} & \multicolumn{1}{c}{R} & \multicolumn{1}{c @{\hspace{0.3in}}}{F1} &  \multicolumn{1}{c @{\hspace{0.3in}}}{F1} \\\midrule

\textbf{1} &	KK2018 &	\textbf{84.3} &	76.0 &	\textbf{79.9} &	65.2 &	73.1 &	\textbf{68.9}&	\textbf{78.5} &	75.4 &	\textbf{76.9} &	\textbf{75.0} \\
 \textbf{2} &	Genius1237 &	81.0 &	\textbf{77.8} &	79.3 &	\textbf{65.7} &	64.3 &	65.0 &	72.0 &	\textbf{77.0} &	74.4 &	72.6 \\
\textbf{3} &	olenet &	78.2 &	74.4 &	76.2 &	62.8 &	65.3 &	64.0 &	75.2 &	75.7 &	75.5 &	71.5	  \\
\textbf{4} &	gopalanvinay &	80.7 &	74.6 &	77.5 &	61.4 &	67.5 &	64.3 &	74.5 &	71.6 &	73.0 &	71.3	\\
\textbf{5} &	ayushk &	78.8 &	73.8 &	76.2 &	60.9 &	67.5 &	64.0 &	75.3 &	70.6 &	72.9 &	70.7 \\ \addlinespace[0.15cm] \hdashline \addlinespace[0.15cm]
 \textbf{6} &	Taha &	78.6 &	72.8 &	75.6 &	60.6 &	70.1 &	65.0 &	76.2 &	67.9 &	71.8 &	70.6\\
 \textbf{7} &	Miriam &	78.0 &	77.3 &	77.6 &	62.6 &	60.1 &	61.3 &	70.7 &	74.9 &	72.7 &	70.2\\
\textbf{8} &	HugoLerogeron &	79.2 &	74.7 &	76.9 &	60.3 &	63.9 &	62.1 &	70.6 &	70.0 &	70.3 &	69.5\\
\textbf{9} &	somban &	78.6 &	72.9 &	75.6 &	59.4 &	65.0 &	62.1 &	71.8 &	69.3 &	70.5 &	69.1\\
\textbf{10} &	adity\_malte &	80.3 &	69.0 &	74.2 &	57.0 &	\textbf{73.5} &	64.2 &	77.3 &	62.2 &	69.0 &	69.0\\ \addlinespace[0.15cm] \hdashline \addlinespace[0.15cm]
\textbf{11} &	MeisterMorxrc &	79.9 &	70.1 &	74.7 &	59.5 &	65.0 &	62.1 &	70.2 &	71.9 &	71.0 &	69.0\\
\textbf{12} &	nirantk &	78.9 &	70.8 &	74.6 &	58.3 &	67.4 &	62.5 &	73.2 &	67.6 &	70.2 &	68.9 \\
\textbf{13} &	apurva19 &	78.8 &	75.8 &	77.3 &	61.2 &	60.8 &	61.0 &	67.4 &	70.8 &	69.1 &	68.8\\
\textbf{14} &	c1pher &	79.7 &	69.7 &	74.4 &	56.5 &	\textbf{73.5} &	63.9 &	78.3 &	60.7 &	68.4 &	68.7\\
\textbf{15} &	will\_go &	77.2 &	70.5 &	73.7 &	57.8 &	70.2 &	63.4 &	75.9 &	63.4 &	69.1 &	68.6\\ \midrule

45 &	\textbf{Baseline} &	72.8 &	68.8 &	70.7 &	56.2 &	60.2 &	58.1 &	69.1 &	67.4 &	68.3 &	65.4\\
 \bottomrule

\end{tabular}}

\caption{Top 15 Results for the \textbf{Hinglish} dataset. 
The systems are ordered by the \textit{Weighted Average F1} (Avg.) scores of the \textit{Postive}, \textit{Neutral}, and \textit{Negative} classes.
We report Precision (P), Recall (R), and F1 score for each class separately.
In each column, the boldfaced scores are the highest score in that column.
}
\label{tab:hinglish_results}
\end{table}

\begin{table}[!htbp]
\centering
\resizebox{0.95\textwidth}{!}{

\begin{tabular}{cl  rrr @{\hspace{0.3in}} rrr  @{\hspace{0.3in}} rrr  @{\hspace{0.3in}} l}
\toprule

\textbf{Rank} & \textbf{System} & 
\multicolumn{3}{c @{\hspace{0.3in}}}{\textbf{Positive}} & 
\multicolumn{3}{c @{\hspace{0.3in}}}{\textbf{Neutral}} & 
\multicolumn{3}{c @{\hspace{0.3in}}}{\textbf{Negative}} & 
\textbf{Avg.} \\ \addlinespace[0.2cm]

& & \multicolumn{1}{c}{P} & \multicolumn{1}{c}{R} & \multicolumn{1}{c @{\hspace{0.3in}}}{F1} & 
\multicolumn{1}{c}{P} & \multicolumn{1}{c}{R} & \multicolumn{1}{c @{\hspace{0.3in}}}{F1} & 
\multicolumn{1}{c}{P} & \multicolumn{1}{c}{R} & \multicolumn{1}{c @{\hspace{0.3in}}}{F1} &  \multicolumn{1}{c @{\hspace{0.3in}}}{F1} \\\midrule

\textbf{1} &	LiangZhao &	88.3 &	92.6 &	\textbf{90.4} &	18.1 &	20.9 &	19.4 &	59.9 &	39.5 &	47.6 &	\textbf{80.6} \\ 
\textbf{2} &	rachel &	89.0 &	87.7 &	88.3 &	16.0 &	45.1 &	23.7 &	\textbf{65.3} &	24.5 &	35.7 &	77.6 \\ 
\textbf{3} &	asking28 &	84.5 &	90.1 &	87.2 &	6.1 &	4.9 &	5.4 &	43.5 &	29.9 &	35.4 &	75.6 \\ 
\textbf{4} &	dpalominop &	91.6 &	77.2 &	83.8 &	12.7 &	30.6 &	17.9 &	42.9 &	5\textbf{8.6} &	49.5 &	75.5 \\ 
\textbf{5} &	kongjun &	87.1 &	84.6 &	85.9 &	11.1 &	30.1 &	16.2 &	56.1 &	27.2 &	36.6 &	75.3 \\ \addlinespace[0.15cm] \hdashline \addlinespace[0.15cm]
\textbf{6} &	HaoYu &	92.9 &	74.0 &	82.4 &	11.9 &	48.1 &	19.1 &	55.2 &	55.0 &	55.1 &	75.2 \\ 
\textbf{7} &	Taha &	84.7 &	89.5 &	87.0 &	\textbf{51.9} &	20.5 &	\textbf{29.4} &	10.4 &	17.5 &	13.0 &	75.1 \\ 
\textbf{8} &	meiyim &	93.0 &	73.3 &	82.0 &	12.1 &	\textbf{55.8} &	19.9 &	57.7 &	47.1 &	51.9 &	74.5 \\ 
\textbf{9} &	Lavinia\_Ap &	82.0 &	97.9 &	89.2 &	13.8 &	3.9 &	6.1 &	56.0 &	8.0 &	14.1 &	74.4 \\ 
\textbf{10} &	jupitter &	\textbf{93.6} &	71.8 &	81.3 &	11.0 &	53.9 &	18.2 &	58.1 &	47.9 &	\textbf{52.5} &	73.9 \\ \addlinespace[0.15cm] \hdashline \addlinespace[0.15cm]
\textbf{11} &	tangmen &	91.8 &	72.5 &	81.0 &	11.3 &	55.3 &	18.8 &	59.8 &	41.6 &	49.0 &	73.0 \\ 
\textbf{12} &	hermosillo748 &	85.4 &	81.3 &	83.3 &	7.3 &	21.8 &	10.9 &	54.1 &	26.6 &	35.7 &	72.8 \\ 
\textbf{13} &	harsh\_6 &	87.7 &	77.9 &	82.5 &	9.5 &	23.3 &	13.5 &	36.1 &	39.1 &	37.5 &	72.5 \\ 
\textbf{14} &	francesita &	80.9 &	\textbf{99.5} &	89.2 &	8.7 &	1.0 &	1.7 &	0.0 &	0.0 &	0.0 &	72.2 \\ 
\textbf{15} &	ajason08 &	90.1 &	71.0 &	79.4 &	8.2 &	40.3 &	13.6 &	54.7 &	37.5 &	44.5 &	71.0  \\ \midrule

23 & \textbf{Baseline} & 89.5 &	63.0 &	74.0 &	7.9 &	49.5 &	13.6 &	47.0 &	31.0 &	37.4 &	65.6  \\ \bottomrule
\end{tabular}}

\caption{Top 15 results for the \textbf{Spanglish} dataset. 
The systems are ordered by the \textit{Weighted Average F1} (Avg.) scores of the \textit{Postive}, \textit{Neutral}, and \textit{Negative} classes.
We report Precision (P), Recall (R), and F1 score for each class separately.
In each column, the boldfaced score is the highest score in that column.
}
\label{tab:spanglish_results}
\end{table}

\begin{table}[!htbp]

\resizebox{0.85\textwidth}{!}{ 
\centering
\begin{adjustbox}{angle=90}

\begin{tabular}{|l|l|l|l|l|l|l|l|l|l|l|l|l|l|l|l|l|} 
\hline
                                    & \multicolumn{4}{l|}{\textbf{~ ~ ~ ~ ~ ~ ~ ~ ~ ~ ~ ~ Embeddings }}            & \multicolumn{8}{l|}{\textbf{~ ~ ~ ~ ~ ~ ~ ~ ~ ~ ~ ~ ~ ~ ~ ~ ~ ~ ~ ~ ~ ~ ML Models }}                                                  & \multicolumn{4}{l|}{\textbf{~ ~ ~ ~ ~ ~ ~ ~ ~ ~ ~LMs }}           \\ 
\hline
                                    & \textbf{Word2Vec} & \textbf{Glove} & \textbf{Character level} & \textbf{fastText} & \textbf{NB} & \textbf{LR} & \textbf{RF} & \textbf{SVM} & \textbf{MLP} & \textbf{CNN} & \textbf{Ensemble} & \textbf{RNN, LSTM and GRU} & \textbf{BERT} & \textbf{XLmR} & \textbf{ALBERT} & \textbf{XLNet}  \\ 
\hline
\textbf{BAKSA}                      &                   &                &                     &                   &             &             &             &              &              & \textbf{\checkmark}   &                   & \textbf{\checkmark}                 & \textbf{\checkmark}    & \textbf{\checkmark}    &                 &                 \\ \hline

\textbf{C1}                         &                   & \textbf{\checkmark}     &                     &                   &             & \textbf{\checkmark}  & \textbf{\checkmark}  & \textbf{\checkmark}   &              &              &                   &                            & \textbf{\checkmark}    &               &                 &                 \\ \hline

\textbf{CS-Embed}        & \textbf{\checkmark}        &                &                     &                   &             &             &             &              &              &              &                   & \textbf{\checkmark}                 &               &               &                 &                 \\ \hline

\textbf{Deep Learning Brasil - NLP} &                   &                &                     &                   &             &             &             &              &              &              & \textbf{\checkmark}        &                            & \textbf{\checkmark}    & \textbf{\checkmark}    & \textbf{\checkmark}      & \textbf{\checkmark}      \\ \hline

\textbf{FII-UAIC}                  &                   &                &                     &                   &             &             &             &              &              & \textbf{\checkmark}   &                   &                            &               &               &                 &                 \\ \hline

\textbf{FiSSA}                      &                   &                &                     &                   &             &             &             &              &              &              &                   & \textbf{\checkmark}                 & \textbf{\checkmark}    & \textbf{\checkmark}    &                 &                 \\ \hline

\textbf{gundapusunil}                    & \textbf{\checkmark}        & \textbf{\checkmark}     & \textbf{\checkmark}          & \textbf{\checkmark}        &             & \textbf{\checkmark}  & \textbf{\checkmark}  & \textbf{\checkmark}   & \textbf{\checkmark}   &              &                   &                            &               &               &                 &                 \\ \hline

\textbf{HCMS}                       &                   &                & \textbf{\checkmark}          & \textbf{\checkmark}        &             &             &             & \textbf{\checkmark}   &              & \textbf{\checkmark}   &                   & \textbf{\checkmark}                 & \textbf{\checkmark}    &               &                 &                 \\ \hline

\textbf{HPCC-YNU}                   &                   &                & \textbf{\checkmark}          &                   &             &             &             &              &              &              &                   & \textbf{\checkmark}                 &               &               &                 &                 \\ \hline

\textbf{HinglishNLP}                    &                   &                &                     &                   & \textbf{\checkmark}  &             &             & \textbf{\checkmark}   &              &              &                   &                            & \textbf{\checkmark}    &               &                 &                 \\ \hline

\textbf{IIITG-ADBU}                 &                   &                &                     &                   &             &             &             & \textbf{\checkmark}   &              &              &                   &                            & \textbf{\checkmark}    &               &                 &                 \\ \hline

\textbf{IIT Gandhinagar}            &                   & \textbf{\checkmark}     &                     &                   &             &             &             &              &              &              &                   & \textbf{\checkmark}                 &               &               &                 &                 \\ \hline

\textbf{IRLab\_DAIICT}               & \textbf{\checkmark}        &                &                     &                   &             & \textbf{\checkmark}  & \textbf{\checkmark}  &              &              &              & \textbf{\checkmark}        &                            & \textbf{\checkmark}    &               &                 &                 \\ \hline

\textbf{IUST}                       & \textbf{\checkmark}        & \textbf{\checkmark}     &                     & \textbf{\checkmark}        &             & \textbf{\checkmark}  &             & \textbf{\checkmark}   &              & \textbf{\checkmark}   & \textbf{\checkmark}        & \textbf{\checkmark}                 & \textbf{\checkmark}    &               &                 &                 \\ \hline

\textbf{JUNLP}                         &                   &                &                     &                   &             &             &             & \textbf{\checkmark}   &              &              &                   &                            &               &               &                 &                 \\ \hline

\textbf{KK2018}                     &                   &                &                     &                   &             &             &             &              &              &              &                   &                            & \textbf{\checkmark}    & \textbf{\checkmark}    &                 &                 \\ \hline

\textbf{LIMSI\_UPV}                  &                   &                &                     & \textbf{\checkmark}        &             &             &             &              &              &              &                   &                            &               &               &                 &                 \\ \hline

\textbf{LT3}                        &                   &                &                     & \textbf{\checkmark}        &             &             &             & \textbf{\checkmark}   &              & \textbf{\checkmark}   &                   & \textbf{\checkmark}                 &               &               &                 &                 \\ \hline

\textbf{MSR India}                  & \textbf{\checkmark}        &                &                     &                   &             &             &             &              &              &              &                   & \textbf{\checkmark}                 & \textbf{\checkmark}    & \textbf{\checkmark}    &                 &                 \\ \hline

\textbf{MeisterMorxrc}              &                   &                &                     &                   &             &             &             &              &              &              &                   &                            & \textbf{\checkmark}    &               &                 &                 \\ \hline

\textbf{NITS-Hinglish-SentiMix}              &                   &                &                     &                   &             &             &             &              &              & \textbf{\checkmark}   & \textbf{\checkmark}        & \textbf{\checkmark}                 &               &               &                 &                 \\ \hline

\textbf{NLP-CIC}                    & \textbf{\checkmark}        &                &                     &                   &             &             &             &              &              & \textbf{\checkmark}   &                   & \textbf{\checkmark}                 &               &               &                 &                 \\ \hline

\textbf{Palomino-Ochoa}                  &                   &                &                     &                   &             &             &             &              &              &              &                   &                            & \textbf{\checkmark}    &               &                 &                 \\ \hline

\textbf{Reed}                       & \textbf{\checkmark}        &                &                     &                   &             &             &             &              &              &              &                   &                            & \textbf{\checkmark}    &               &                 &                 \\ \hline

\textbf{Team\_Swift}                 &                   &                &                     & \textbf{\checkmark}        &             &             &             &              &              &              &                   & \textbf{\checkmark}                 &               & \textbf{\checkmark}    &                 &                 \\ \hline

\textbf{TueMix}                    &                   &                &                     &                   &             & \textbf{\checkmark}  &             &              &              & \textbf{\checkmark}   &                   & \textbf{\checkmark}                 &               &               &                 &                 \\ \hline

\textbf{ULD@NUIG}                      &                   &                &                     &                   &             &             &             & \textbf{\checkmark}   &              & \textbf{\checkmark}   &                   & \textbf{\checkmark}                 &               &               &                 &                 \\ \hline

\textbf{UPB}                        &                   &                &                     &                   &             &             &             & \textbf{\checkmark}   &              & \textbf{\checkmark}   &                   & \textbf{\checkmark}                 & \textbf{\checkmark}    & \textbf{\checkmark}    &                 &                 \\ \hline

\textbf{Voice@SRIB}                 &                   &                &                     &                   &             &             &             &              &              & \textbf{\checkmark}   & \textbf{\checkmark}        & \textbf{\checkmark}                 &               &               &                 &                 \\ \hline

\textbf{WESSA}                      & \textbf{\checkmark}        &                &                     &                   & \textbf{\checkmark}  & \textbf{\checkmark}  &             & \textbf{\checkmark}   &              &              &                   &                            &               & \textbf{\checkmark}    &                 &                 \\ \hline

\textbf{Will\_go}                    & \textbf{\checkmark}        &                & \textbf{\checkmark}          &                   &             & \textbf{\checkmark}  &             &              &              &              & \textbf{\checkmark}        &                            & \textbf{\checkmark}    &               &                 &                 \\ \hline

\textbf{XLP}                        &                   &                &                     &                   &             &             &             &              &              & \textbf{\checkmark}   &                   &                            & \textbf{\checkmark}    & \textbf{\checkmark}    &                 &                 \\ \hline

\textbf{Zyy1510}                    &                   &                & \textbf{\checkmark}          &                   & \textbf{\checkmark}  &             &             & \textbf{\checkmark}   &              &              & \textbf{\checkmark}        & \textbf{\checkmark}                 &               &               &                 &                 \\ \hline

\end{tabular}
\end{adjustbox}
}
\caption{Overview of the techniques used by the participants. This is not an exhaustive list. Teams are sorted alphabetically.  }

\label{tab:teams_techniques}
\end{table}

In the previous section, we briefly described the top systems. Here, we group and summarize various techniques used by the systems (Codalab usernames are mentioned in parentheses) : 

\begin{itemize}

\item \textbf{Word Embedding:} Three popular word embedding ways explored by participants. Word2Vec, Glove, FastText. Some participants used character-embedding. Additional resources were also used by participants to train their own embeddings.

\item \textbf{Classical ML methods:} Classical ML techniques like - logistic regression, Naive Bayes, Perceptron, and SVM have been tested by several researchers. Naive Bayes and its multinomial kernel was tried by Zyy1510 (zyy1510). Teams like TueMix (guzimanis), WESSA (ahmed0sultan), C1 (lakshadvani) reported their experiments with Logistic Regression, whereas yet another popular choice Random Forest has been used by IRLab\_DAIICT (apurva19), C1 (lakshadvani). SVM was tried by quite a few teams - IUST (Taha), JUNLP (sainik.mahata), WESSA (ahmed0sultan), C1 (lakshadvani), IIITG-ADBU (abaruah).


\item \textbf{RNN:} RNN, GRU, LSTM, along with their bi-directional varients were explored by several teams. Some of them are gundapusunil (gundapusinil), Team\_Swift (aditya\_malte), CS-Embed (francesita), GULD@NUIG (koustava), IIT Gandhinagar (vivek\_IITGN) etc.

\item \textbf{CNN for text:} Although RNN is the more popular choices for NLP tasks, quite a few teams also used CNN for text. Some of them are IUST (Taha), FII-UAIC (Lavinia\_AP), NLP-CIC (ajason08), NITS-Hinglish-SentiMix (rns2020), Zyy1510 (zyy1510), HCMS (the0ne, talent404).

\item \textbf{Transformer, BERT and related language models:} The recent trend in NLP is to use highly capable language models like BERT and XLNet. The popular choice, BERT, was tried by MeisterMorxrc (MeisterMorxrc), HinglishNLP (nirantk), IRLab\_DAIICT (apurva19), WESSA (ahmed0sultan), C1 (lakshadvani), IIITG-ADBU (abaruah). Some researchers like Deep Learning Brasil - NLP (verissimo.manoel) experimented with XLNet. XLmR was used by Will\_go (will\_go) , kk2018 (kk2018), FiSSA (jupitter) etc.  These type of models gave the best results.

\item \textbf{Ensembles:} Some teams like Voice@SRIB (asking28), UPB (eduardgzaharia, clementincercel) etc. used ensemble methods. in all cases, ensembles performed better than the their individual models. 

\item \textbf{Special Mentions:} Apart from common practices and architectures quite a few participants explored interesting dimensions and added significant value to this endeavor. We strongly believe these dimensions need to be explored and discussed further.:

\textbf{XLP (LiangZhao)} used Cross-lingual embeddings which could an interesting way for code-mixed language processing where we have scarcity of annotated data. 

 \textbf{UPB (eduardgzaharia, clementincercel)} used capsule network with biGRU and showed promising results. The use of capsule networks in NLP tasks need further exploration.

\textbf{ULD@NUIG (koustava)} explored an interesting way to phoneme based Generative Morphemes learning approach. Sub-word based embedding is an interesting new way in the NLP community, but what is the best sub-word unit to choose is still unresolved. Morpheme based approach could be a good alternative, especially for highly spelling variant code-mixed data. 

\textbf{IIT Gandhinagar (vivek\_IITGN)} tried a new direction by generating sentences using language modeling. Language modeling for code-mixed data is still an under-researched problem.

\textbf{ HPCC-YNU (kongjun)} used a Bilingual Vector Gating Mechanism. Vector gating technique got certain success in document classification kinds of applications, but its applications in other NLP dimension demands further exploration.

\textbf{Will\_go (will\_go)} used Bert and Pseudo labeling. Pseudo Labeling can be a useful strategy for code-mixed languages especially when annotated data is scarce. .

\textbf{kk2018 (kk2018)} reported unique ways to apply adversarial network and its usage in code-mixing. They got very good results. 

\textbf{LIMSI\_UPV (somban)} gave a way to merge RNN and CNN architecture together for the betterment of sentiment analysis. This could be an interesting way to explore in the future. 

\end{itemize}

\section{Conclusion and Future Work}\label{sec:conclusion}
SentiMix, sentiment analysis of code-mixed tweets at SemEval 2020 received an overwhelming response for both Hinglish and Spanglish. 61 teams submitted their systems for Hinglish and 28 teams submitted their systems for Spanglish. The best performance achieved was 75.0 \% F1 score for Hinglish and 80.6\% for Spanglish. We received a total of 33 system description papers. BERT-like models were the most successful among participants. Although the SentiMix task mainly focused on sentiment analysis, the data will serve the NLP community or whoever is interested in  the  code-mixing problem for these particular languages and in general. 

Properly annotated code-mixed data is still scarce. The success of SentiMix motivates us to go further and organize similar events in the future. We plan to add more languages, especially from regions that have a high percentage of bi- or multilingual speakers. We also plan to enrich our datasets with annotations for other tasks (NER, emotion recognition, translation etc). We strongly believe that code-mixing is a new horizon of interest in the NLP community and needs to be further explored in the future.


\bibliographystyle{acl}
\bibliography{coling2020}

\newpage

\appendix

\section{Participants}

\begin{table}[!hbtp]
    \centering
    \begin{tabular}{ll l}
    \toprule
    
    \textbf{Team} & \textbf{Codalab Usernames} & \textbf{System Description Paper}\\ \midrule
    
    BAKSA           & ayushk, harsh\_6        & \newcite{BAKSA}\\
    C1              & lakshadvani   & \newcite{C1}\\
    CS-Embed        & francesita    & \newcite{CS-Embed} \\
    Deep Learning Brasil - NLP & verissimo.manoel & \newcite{DeepLearningBrasil} \\
    FII-UAIC        & Lavinia\_Ap   & \newcite{FII-UAIC} \\

    FiSSA           & jupitter      & \newcite{FiSSA} \\
    gundapusunil    & gundapusunil  & \newcite{gundapusunil} \\
    HCMS            & the0ne, talent404        & \newcite{HCMS}\\
    HPCC-YNU        & kongjun       & \newcite{HPCC-YNU}\\
    HinglishNLP     & Nirantk & \newcite{HinglishNLP} \\ 
    IIITG-ADBU      & abaruah       & \newcite{IIITG-ADBU} \\

    IIT Gandhinagar & vivek\_IITGN  & \newcite{IIT} \\
    IRLab\_DAIICT   & apurva19      & \newcite{IRLab}\\
    IUST            & Taha          & \newcite{IUST}\\
    JUNLP           & sainik.mahata & \newcite{JUNLP}\\
    KK2018          & kk2018        & \newcite{kk2018} \\
    
    LIMSI\_UPV       & somban        & \newcite{LIMSI}\\
    LT3             & c1pher        & \newcite{LT3}\\
    MSR India       & genius1237    & \newcite{MSR} \\
    MeisterMorxrc   & MeisterMorxrc & \newcite{MeisterMorxrc}\\
    NITS-Hinglish-SentiMix   & rns2020       & \newcite{NITS} \\
    
    NLP-CIC         & ajason08      & \newcite{NLP-CIC} \\
    Palomino-Ochoa  & dpalominop    & \newcite{Palomino-Ochoa}\\
    Reed            & gopalvinay    & \newcite{Reed}\\
    Team\_Swift      & aditya\_malte & \newcite{Team_Swif}\\
    TueMix          & guzimanis     & \newcite{TueMix} \\
    
    ULD@NUIG        & koustava      & \newcite{ULD@NUIG}\\
    UPB             & eduardgzaharia, clementincercel & \newcite{UPB} \\
    Voice@SRIB      & asking28      & \newcite{Voice@SRIB}\\
    WESSA           & ahmed0sultan  & \newcite{WESSA} \\
    Will\_go         & will\_go      & \newcite{Will}\\
    XLP             & LiangZhao     & \newcite{XLP}\\
    Zyy1510         & zyy1510       & \newcite{Zyy1510} \\

     \\

    \hline
    \end{tabular}
    \caption{The teams that participated in Sentimix-2020 and submitted system description papers with the corresponding reference thereof. Teams are sorted alphabetically. }
    \label{tab:participants_bib}
\end{table}

\end{document}